\documentclass[11pt]{article}
\usepackage[preprint]{neurips_2025}
\usepackage[utf8]{inputenc}
\usepackage[T1]{fontenc}
\usepackage{hyperref}
\usepackage{url}
\usepackage{booktabs}
\usepackage{amsfonts}
\usepackage{amsmath}
\usepackage{amssymb}
\usepackage{nicefrac}
\usepackage{microtype}
\usepackage{graphicx}
\usepackage{multirow}
\usepackage{enumitem}

\title{Mitigating Shortcut Reasoning in Language Models: A Gradient-Aware Training Approach}

\author{
Hongyu Cao \\
Arizona State University \\
\texttt{hongyuca@asu.edu}
\And
Kunpeng Liu \\
Clemson University \\
\texttt{kunpenl@clemson.edu}
\And
Dongjie Wang \\
University of Kansas \\
\texttt{wangdongjie100@gmail.com}
\And
Yanjie Fu \\
Arizona State University \\
\texttt{yanjiefu@asu.edu}
}

\begin{document}
\maketitle
\begin{abstract}
Large language models exhibit strong reasoning capabilities, yet often rely on shortcuts—surface pattern matching, memorization, and keyword correlations—rather than genuine logical inference. This stems from training signal misalignment: shortcut-encouraging examples reduce loss but impair generalization when distributions shift or problem phrasing varies.
We formulate \textbf{shortcut-aware reasoning training}: identifying samples that promote shortcut updates and modifying gradient dynamics to emphasize generalizable reasoning. A key challenge is that shortcut samples efficiently reduce training loss, making them difficult to detect via conventional metrics. Existing approaches—Chain-of-Thought training, Self-Consistency decoding, data filtering, and RLHF—fail because they do not modify training dynamics or detect shortcut-promoting gradients.
We identify a \textbf{new opportunity}: shortcut reasoning samples produce distinct gradient signatures, exhibiting low alignment with validation-improving gradients and high concentration on answer tokens. These signals enable detection and mitigation through gradient structure analysis rather than data labels alone.
We propose a method combining \textbf{ShortcutScore} for sample reweighting with \textbf{gradient surgery techniques}. Our pipeline integrates periodic validation gradient computation with empirical validation across diverse datasets. Experiments demonstrate significant improvements in robustness and reasoning reliability, enhancing LLM generalization toward genuine reasoning. Code available at: \url{https://github.com/fuyanjie/short-cut-aware-data-centric-reasoning}
\end{abstract}
\section{Introduction}

Large language models (LLMs) deployed in high-stakes reasoning tasks frequently exploit \textit{reasoning shortcuts}—surface pattern matching, answer memorization, keyword-answer correlations, and premature answer prediction—rather than performing genuine logical inference~\cite{wei2022chain,wang2022self}. This shortcut reliance stems from \textbf{training signal misalignment}: shortcut-encouraging samples efficiently reduce training loss but fail to foster generalizable reasoning. Consequently, models accurate on training distributions fail catastrophically when numbers change, constraints vary, or problem phrasing shifts. This motivates \textit{\textbf{S}hortcut-\textbf{A}ware \textbf{R}easoning \textbf{T}raining (SART)}: identifying samples that promote shortcut updates and modifying gradient dynamics to emphasize genuine reasoning signals. Solving SART is critical for reliable LLM deployment in mathematical reasoning, financial analysis, planning, and scientific discovery—domains where spurious reasoning leads to incorrect decisions or unsafe recommendations.

Two major challenges arise in solving SART: (1) detecting shortcut-promoting samples from gradient behavior, and (2) modifying training dynamics to suppress shortcut learning without degrading performance. First, shortcut samples are often correctly labeled and indistinguishable by conventional quality metrics. They produce strong gradients that efficiently reduce training loss, making loss-based detection unreliable. The detection challenge is: how can we identify training samples whose gradients improve loss but harm reasoning generalization? Second, simply down-weighting or removing detected shortcut samples is insufficient—it destabilizes training and discards useful signal. The suppression challenge is: how can we modify training dynamics to neutralize shortcut gradients while preserving generalizable reasoning signals from the same samples?

Existing methods only partially address SART. Chain-of-Thought (CoT) training~\cite{wei2022chain,kojima2022large} encourages intermediate reasoning traces but does not modify training dynamics: models can generate plausible steps that still rely on shortcuts, and shortcut gradients remain dominant~\cite{ho2022large}. Self-Consistency Decoding~\cite{wang2022self} samples multiple reasoning paths to detect inconsistencies but operates at inference time, leaving shortcut patterns intact in model parameters. Data filtering and curriculum learning~\cite{bengio2009curriculum} attempt to remove noisy samples, but shortcut samples are typically correctly labeled, making them invisible to label-quality filters. RLHF~\cite{ouyang2022training} optimizes reward signals based on answer correctness rather than reasoning validity, allowing shortcuts to persist when they produce correct answers. These approaches cannot jointly address shortcut detection from gradient behavior and dynamic suppression during training—a gradient-centric perspective is required.

\textbf{Our insights: gradient-aware shortcut detection and correction.} We formulate SART as a gradient structure analysis problem. We find that shortcut reasoning samples produce two distinct gradient signatures: (1) \textit{low alignment} with gradients that improve held-out validation performance, indicating the sample's gradient direction does not transfer to generalizable reasoning; and (2) \textit{high answer-token concentration}, where gradient norms are dominated by final answer tokens rather than intermediate reasoning tokens. These signals—gradient alignment with validation gradients and answer-gradient concentration—can be combined into a principled \textbf{ShortcutScore} quantifying the degree to which a sample promotes shortcut learning. Furthermore, harmful shortcut gradient directions can be neutralized via \textit{gradient surgery}: projecting gradients orthogonally onto subspaces that avoid non-transferable directions, forcing the model to learn from genuine reasoning signals without discarding samples entirely.

\textbf{Proposed Approach.} This paper presents the first principled shortcut-aware training framework for LLM reasoning via gradient structure analysis, with two goals: (1) precise detection of shortcut-promoting samples through gradient behavior; (2) effective suppression of shortcut learning while preserving generalizable reasoning signals.
For Goal 1, we develop the \textbf{ShortcutScore}, a composite metric combining cosine similarity between per-sample gradients and validation gradients (measuring non-transfer alignment) with the ratio of answer-token to reasoning-token gradient norms (measuring answer-dominant credit assignment). Samples with high ShortcutScore are flagged as shortcut-promoting.
For Goal 2, we develop a \textbf{gradient surgery} technique that identifies harmful gradient directions—those misaligned with validation gradients or increasing ShortcutScore—and projects them orthogonally to suppress their influence during parameter updates. This ensures training dynamics prioritize reasoning token signals.

Our framework is model-agnostic and integrates into any LLM training pipeline without architectural changes. Extensive experiments across mathematical reasoning, commonsense reasoning, and constraint-based planning benchmarks demonstrate significant improvements in robustness under distribution shift, number perturbation, and phrasing variation, validating that our approach reduces shortcut reliance and promotes genuine logical inference. Code is available at \url{https://github.com/fuyanjie/short-cut-aware-data-centric-reasoning}.
\section{Problem Statement}

\textbf{The SART Problem.}
We study supervised training of language models on reasoning tasks. Let $\mathcal{D} = \{(x_i, y_i)\}_{i=1}^{N}$ be a training dataset where $x_i$ is an input reasoning problem and $y_i$ is the corresponding output comprising intermediate reasoning steps and a final answer. Let $\mathcal{V} = \{(x_j, y_j)\}_{j=1}^{M}$ be a small held-out validation set drawn from a distribution that rewards genuine reasoning, and let $\theta$ denote the parameters of a language model $f_\theta$. Standard training minimizes the empirical risk:
\begin{equation}
    \min_{\theta} \; \frac{1}{N} \sum_{i=1}^{N} \ell(\theta;\, x_i, y_i),
\end{equation}
where $\ell$ is the token-level negative log-likelihood loss. This objective does not distinguish between samples that promote genuine logical inference and those that achieve low loss via superficial patterns---keyword-answer correlations, answer token memorization, or template matching. We call such samples \emph{shortcut reasoning samples}.

Formally, let $g_s = \nabla_\theta\, \ell(\theta; s)$ denote the gradient of a training sample $s = (x, y)$, and let
\begin{equation}
    g_{\mathcal{V}} = \nabla_\theta \; \frac{1}{|\mathcal{V}|} \sum_{v \in \mathcal{V}} \ell(\theta;\, v)
\end{equation}
be the validation gradient, reflecting parameter updates that improve generalizable reasoning. A sample $s$ is a shortcut reasoning sample if its gradient $g_s$ (i) exhibits low alignment with $g_{\mathcal{V}}$, indicating the update does not transfer to genuine reasoning, and (ii) concentrates gradient norm on final answer tokens rather than intermediate reasoning tokens, indicating the model learns \emph{what to answer} rather than \emph{how to reason}. Critically, such samples are often correctly labeled and efficiently reduce training loss, rendering them undetectable by conventional data quality or label-noise metrics.

\textbf{Shortcut-Aware Reasoning Training (SART)} addresses this by learning parameters $\theta$ that minimize training loss while suppressing shortcut-promoting gradient directions:
\begin{equation}
    \min_{\theta} \; \frac{1}{N} \sum_{i=1}^{N} w_i(\theta)\; \ell(\theta;\, x_i, y_i),
\end{equation}
where $w_i(\theta) \in [0,1]$ downweights samples that promote shortcut reasoning, subject to gradient updates remaining aligned with $g_{\mathcal{V}}$. SART pursues two goals: (1) \emph{detect} shortcut samples via gradient structure---specifically, non-transfer alignment and answer-token concentration; and (2) \emph{suppress} shortcut learning by neutralizing harmful gradient directions while preserving generalizable reasoning signals within the same samples. The framework operates purely at the level of training dynamics, requiring neither architectural modifications nor additional human annotations.
\section{Overview}

\begin{figure}
    \centering
    \includegraphics[width=\linewidth]{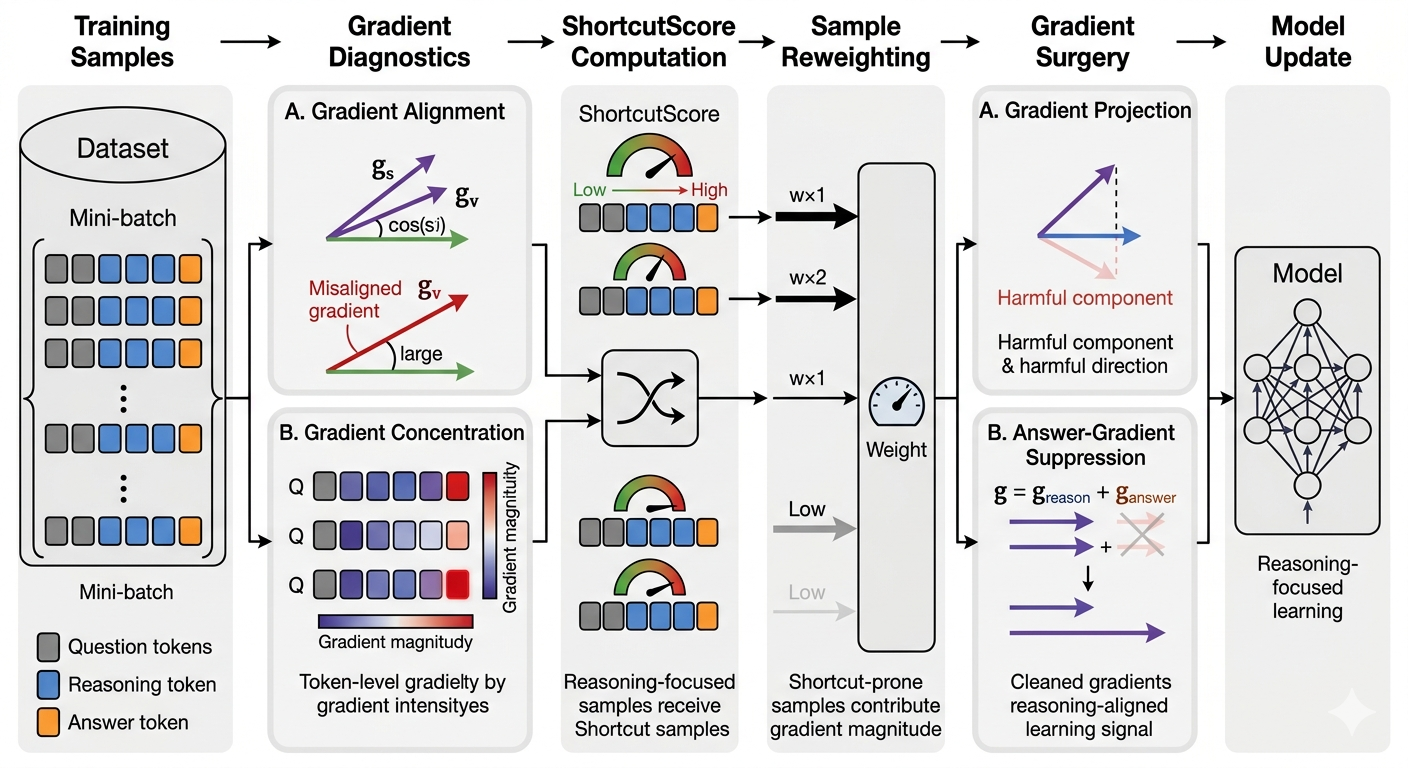}
    \caption{
    Overview of SART. (A) Gradient diagnostics: we compute per-sample gradient alignment with validation gradients and answer-token concentration. 
    (B) ShortcutScore computation combines both signals to identify shortcut-prone samples. 
    (C) Sample reweighting reduces their contribution. 
    (D) Gradient surgery projects harmful components out of updates. 
    (E) Resulting updates prioritize reasoning signals over shortcut patterns.
    }
    \label{fig:overview}
\end{figure}
Figure~\ref{fig:overview} illustrates our proposed \textbf{shortcut-aware reasoning training (SART)} framework, comprising two coupled components: (1) \textbf{ShortcutScore-based sample reweighting} and (2) \textbf{gradient surgery for shortcut suppression}.

\textbf{Shortcut Detection via ShortcutScore.}
We identify shortcut reasoning samples by analyzing gradient structure rather than labels or loss values. Signal 1 measures \textit{non-transfer gradient alignment}: we compute the cosine similarity between each sample's per-sample gradient and the validation gradient $g_{\mathcal{V}}$. Samples whose gradients are misaligned with $g_{\mathcal{V}}$ reduce training loss without improving generalizable reasoning---a hallmark of shortcut exploitation undetectable by label-quality or loss-based filters. Signal 2 measures \textit{answer-gradient concentration}: we compare gradient norm on final answer tokens to that on intermediate reasoning tokens. Shortcut samples concentrate gradient energy on answer tokens, indicating the model learns \textit{what to predict} rather than \textit{how to reason}. We combine both signals into a unified \textbf{ShortcutScore} that quantifies each sample's shortcut tendency and down-weights it during training.

\textbf{Shortcut Suppression via Gradient Surgery.}
Reweighting reduces shortcut influence but does not eliminate the harmful gradient directions such samples introduce. We therefore project the gradients of high-ShortcutScore samples onto subspaces that avoid directions which increase the ShortcutScore or decrease alignment with $g_{\mathcal{V}}$, retaining useful reasoning signals while neutralizing shortcut-promoting components. Identifying the correct projection subspace is formulated as a \textbf{minimax optimization problem}: the inner maximization finds the most harmful gradient direction, while the outer minimization ensures updates avoid it. We develop an efficient solver for this minimax game that performs periodic gradient surgery at each update step without significant computational overhead. Together, ShortcutScore reweighting and gradient surgery steer training dynamics toward genuine reasoning signals rather than superficial patterns.
\subsection{ShortcutScore-Based Sample Reweighting}

\noindent\textbf{Why ShortcutScore Reweighting Matters.}
Standard empirical risk minimization treats all training samples equally, optimizing the average token-level loss across the dataset. This is problematic because shortcut reasoning samples---those that achieve low loss by exploiting keyword correlations, answer memorization, or surface pattern matching---generate strong gradients that efficiently reduce training loss but do not support generalizable reasoning. Critically, such samples are often correctly labeled, making them invisible to conventional data quality filters or loss-based selection criteria. The consequence is a systematic training signal misalignment: the model is rewarded for learning spurious strategies, leading to brittleness under distribution shift, number perturbation, or rephrasing. This motivates a reweighting strategy that identifies shortcut-promoting samples from their gradient behavior and reduces their influence on parameter updates---targeting the root cause rather than the symptom.

\noindent\textbf{Modeling Intuition.}
Our central insight is that shortcut reasoning samples produce two distinguishable gradient signatures. First, their per-sample gradients are \textit{poorly aligned} with the validation gradient $g_{\mathcal{V}}$: the update direction they induce does not improve held-out reasoning performance, signaling that the learned signal does not transfer. Second, their gradient energy is \textit{disproportionately concentrated on answer tokens} rather than intermediate reasoning tokens: the model learns \emph{what to predict} rather than \emph{how to reason}. We combine these two signals into a unified \textbf{ShortcutScore} $S(s)$ that quantifies each sample's tendency to promote shortcut learning, and use it to down-weight shortcut-prone samples during optimization.

\noindent\textbf{Step 1: Non-Transfer Gradient Alignment.}
This step measures how well a sample's gradient direction transfers to generalizable reasoning. For each training sample $s = (x, y)$, we compute the per-sample gradient $\mathbf{g}_s = \nabla_\theta \ell(\theta; s)$ and the validation gradient
\begin{equation}
    \mathbf{g}_{\mathcal{V}} = \nabla_\theta \frac{1}{|\mathcal{V}|} \sum_{v \in \mathcal{V}} \ell(\theta; v).
\end{equation}
The non-transfer gradient alignment is then defined as:
\begin{equation}
    A(s) = \cos(\mathbf{g}_s,\, \mathbf{g}_{\mathcal{V}}) = \frac{\mathbf{g}_s^\top \mathbf{g}_{\mathcal{V}}}{\|\mathbf{g}_s\|\|\mathbf{g}_{\mathcal{V}}\|}.
\end{equation}
A low value of $A(s)$ indicates that the gradient of sample $s$ points in a direction inconsistent with improving validation reasoning performance---a hallmark of shortcut exploitation.

\noindent\textbf{Step 2: Answer-Gradient Concentration.}
This step measures how much of the gradient signal is driven by answer tokens versus reasoning tokens. Let $T_{\mathrm{ans}}$ and $T_{\mathrm{reason}}$ denote the index sets of answer tokens and reasoning tokens in the output sequence, respectively. The answer-gradient concentration is defined as:
\begin{equation}
    R(s) = \frac{\sum_{t \in T_{\mathrm{ans}}} \|\nabla_\theta \ell_t\|}{\sum_{t \in T_{\mathrm{ans}} \cup T_{\mathrm{reason}}} \|\nabla_\theta \ell_t\|},
\end{equation}
where $\ell_t$ is the token-level loss at position $t$. A high value of $R(s)$ indicates that the model's parameter updates are driven predominantly by answer prediction rather than by learning intermediate reasoning steps.

\noindent\textbf{Step 3: ShortcutScore and Reweighted Objective.}
We combine the two signals into the \textbf{ShortcutScore}:
\begin{equation}
    S(s) = \alpha \cdot \max(0,\, \tau_A - A(s)) + \beta \cdot \max(0,\, R(s) - \tau_R),
\end{equation}
where $\tau_A$ and $\tau_R$ are thresholds defining low alignment and high concentration, and $\alpha, \beta$ are balancing hyperparameters. Samples with high $S(s)$ are flagged as shortcut-promoting. We assign each sample a weight $w(s) = \exp(-\lambda S(s))$, so that higher shortcut tendency leads to lower training influence. The reweighted optimization objective becomes:
\begin{equation}
    \min_{\theta}\; \frac{1}{N}\sum_{i=1}^{N} w(s_i) \cdot \ell(\theta;\, s_i).
\end{equation}
The model parameters are updated at each step via:
\begin{equation}
    \theta_{t+1} = \theta_t - \eta \sum_{s \in \text{batch}} w(s) \cdot \nabla_\theta \ell(\theta;\, s),
\end{equation}
where $\eta$ is the learning rate. Validation gradients $\mathbf{g}_{\mathcal{V}}$ are recomputed periodically to keep the alignment signal up to date without excessive overhead. A detailed training procedure is provided in Appendix~A.1.

%-----------------------------------------------------------
\subsection{Gradient Surgery for Shortcut Suppression}

\noindent\textbf{Why Gradient Surgery Matters.}
Reweighting reduces the \emph{magnitude} of shortcut-promoting gradients but does not correct their \emph{direction}. Even a down-weighted gradient can still contain a harmful directional component that nudges model parameters away from genuine reasoning. The structural reason is that shortcut samples carry two types of harmful gradient components: (i) directions that are misaligned with validation gradients, and (ii) directions dominated by answer token signals. Simply scaling these gradients down preserves both failure modes. We therefore propose gradient surgery~\cite{yu2020gradient}: an operation that actively projects shortcut gradients onto subspaces that avoid harmful directions, ensuring that only genuine reasoning signals drive parameter updates.

\noindent\textbf{Modeling Intuition.}
The intuition is analogous to the flat region concept in continual learning~\cite{schulman2015trust}: we seek parameter update directions that are simultaneously beneficial for reasoning generalization and minimally disruptive to the reasoning signals already learned. For a shortcut sample $s$, its gradient $\mathbf{g}_s$ can be decomposed into a harmful component (aligned with non-transferable or answer-dominant directions) and a residual component (aligned with genuine reasoning signals). Gradient surgery removes the harmful component and retains the residual, so that the update from sample $s$ contributes constructively to reasoning learning.

\noindent\textbf{Step 1: Projection for Non-Transfer Gradient Alignment.}
For samples with low alignment $A(s) < \tau_A$, we remove the gradient component that lies in the direction of $\mathbf{g}_{\mathcal{V}}$, retaining only the orthogonal complement. Formally:
\begin{equation}
    \mathbf{g}_s' = \mathbf{g}_s - \gamma \frac{\mathbf{g}_s^\top \mathbf{g}_{\mathcal{V}}}{\|\mathbf{g}_{\mathcal{V}}\|^2}\, \mathbf{g}_{\mathcal{V}},
\end{equation}
where $\gamma \in [0,1]$ controls the strength of the projection. Setting $\gamma=1$ fully removes the non-transferable component. The resulting $\mathbf{g}_s'$ is orthogonal to $\mathbf{g}_{\mathcal{V}}$, ensuring the update does not reinforce non-generalizable directions.

\noindent\textbf{Step 2: Suppression for Answer-Gradient Concentration.}
For samples with high concentration $R(s) > \tau_R$, we decompose $\mathbf{g}_s$ into answer-token gradients $\mathbf{g}_s^{\mathrm{ans}}$ and reasoning-token gradients $\mathbf{g}_s^{\mathrm{reason}}$, and suppress the answer-dominant component:
\begin{equation}
    \mathbf{g}_s'' = (1 - \rho)\,\mathbf{g}_s^{\mathrm{ans}} + \mathbf{g}_s^{\mathrm{reason}},
\end{equation}
where $\rho \in [0,1]$ is the suppression coefficient. Setting $\rho=1$ entirely removes the answer token gradient contribution, forcing the model to update based solely on intermediate reasoning signals.

\noindent\textbf{Minimax Formulation for Optimal Surgery.}
A key challenge is determining the optimal surgery direction and strength automatically, rather than relying on fixed thresholds. We find that this challenge can be reformulated as a \textbf{minimax game}: the inner maximization identifies the most harmful gradient direction $\xi$ within a neighborhood of the current parameters, and the outer minimization finds parameter updates that remain robust under this worst-case perturbation:
\begin{equation}
\begin{aligned}
    \min_{\theta}\max_{\xi} \quad & \sum_{s \in \mathcal{D}} w(s)\cdot\ell(\theta + \xi;\, s) \\
    \mathrm{s.t.} \quad & \xi \in \mathcal{M},
\end{aligned}
\end{equation}
where $\mathcal{M}$ is the subspace spanned by previous parameter directions, constraining perturbations to directions relevant to shortcut signals. This formulation automatically identifies both the harmful direction and the appropriate surgery strength, without requiring manual threshold tuning.

\noindent\textbf{Solving the Optimization.}
Based on gradient projection~\cite{rosen1960gradient}, the adversarial perturbation $\xi$ is updated via:
\begin{equation}
    \xi \leftarrow \xi + \eta_1\, \mathrm{proj}_{\mathcal{M}}\!\left(\nabla_\theta \sum_{s \in \mathcal{D}} w(s)\cdot\ell(\theta + \xi;\, s)\right).
\end{equation}
The model parameters $\theta$ are then updated along the direction orthogonal to $\mathcal{M}$, ensuring minimal disruption to previously learned reasoning knowledge:
\begin{equation}
    \theta \leftarrow \theta - \eta_2\,(I - \mathrm{proj}_{\mathcal{M}})\left(\nabla_\theta \sum_{s \in \mathcal{D}} w(s)\cdot\ell(\theta + \xi;\, s)\right),
\end{equation}
where $I$ is the identity matrix and $\eta_2$ is the step size. The complete training pipeline integrates ShortcutScore computation, sample reweighting, and gradient surgery at each update step, with periodic recomputation of $\mathbf{g}_{\mathcal{V}}$. A full algorithmic description is provided in Appendix~A.2.
%=============================================================================
\section{Experiments}
\label{sec:experiments}

We evaluate SART on three synthetic reasoning benchmarks with controlled shortcut injection, comparing against eleven baselines spanning data-centric, loss-based, distributionally robust, and invariant learning approaches.

%-----------------------------------------------------------------------------
\subsection{Experimental Setup}
\label{sec:exp_setup}

\paragraph{Model and Training.}
We use a GPT-style transformer (4 layers, $d_{\mathrm{model}}=256$, 8 attention heads, $d_{\mathrm{ff}}=1024$, ${\approx}$3.2M parameters) trained for 40 epochs with AdamW~\cite{loshchilov2019decoupled} (learning rate $1{\times}10^{-3}$, weight decay $10^{-4}$, batch size 64, cosine annealing). SART hyperparameters ($\alpha{=}\beta{=}1.0$, $\tau_A{=}0.3$, $\tau_R{=}0.5$, $\lambda{=}3.0$, $\gamma{=}1.0$, $\rho{=}0.5$) are selected via grid search over 80 configurations (\S\ref{sec:exp_sensitivity}).

\paragraph{Datasets.}
We construct three synthetic datasets with controlled shortcut injection, each containing 2{,}000 training / 500 validation / 1{,}000 test samples. Training data follows the shortcut rule for 70\% of samples and the true rule for 30\%; validation and test sets exclusively follow the true rule. The test set is split evenly into \emph{clean} (shortcut consistent with correct answer) and \emph{perturbed} (shortcut contradicts the correct answer) subsets.
\begin{itemize}[leftmargin=*, itemsep=1pt]
    \item \textbf{Math-Arithmetic}: Binary classification of $a+b \geq 10$. Shortcut: first operand $a \geq 5$; true rule: evaluate the full sum.
    \item \textbf{Financial-Analysis}: Regulatory compliance from four financial features. Shortcut: revenue $\geq 5$; true rule: margin $\geq 5$ \emph{and} debt $< 5$ (multi-feature conjunction).
    \item \textbf{Causal-Reasoning}: Causal direction inference from observed variables with a confounder. Shortcut: correlation $\geq 5$; true rule: $x \geq 5$ \emph{and} $z < 3$ (requires confounder adjustment).
\end{itemize}

\paragraph{Baselines.}
We compare against eleven methods:
(1)~\textbf{SFT}: vanilla cross-entropy;
(2)~\textbf{Self-Consistency}~\cite{wang2023selfconsistency}: majority vote over 5 samples ($T{=}0.8$);
(3)~\textbf{Data Filtering}: remove high-confidence samples ($> 0.90$) after warmup;
(4)~\textbf{JTT}~\cite{liu2021justtraintwiceimproving}: upweight error-prone samples;
(5)~\textbf{Focal Loss}~\cite{lin2018focallossdenseobject}: $(1{-}p_t)^\gamma$ weighting;
(6)~\textbf{Group DRO}$^\dagger$~\cite{sagawa2020distributionallyrobustneuralnetworks}: worst-group optimization with known shortcut annotations;
(7)~\textbf{IRM}$^\ddagger$~\cite{arjovsky2020invariantriskminimization}: invariance penalty across environments;
(8)~\textbf{V-REx}$^\ddagger$~\cite{krueger2021outofdistributiongeneralizationriskextrapolation}: variance of risks across environments;
(9)~\textbf{Fishr}$^\ddagger$~\cite{rame2022fishrinvariantgradientvariances}: gradient variance equalization;
(10)~\textbf{LfF}~\cite{nam2020learningfailuretrainingdebiased}: learning from failure;
(11)~\textbf{Influence Filtering}~\cite{koh2020understandingblackboxpredictionsinfluence}: remove harmful samples via influence functions.
Here $^\dagger$ requires group annotations and $^\ddagger$ requires multiple environments.

\paragraph{Metrics.}
Clean accuracy, perturbed accuracy (\emph{robustness}), reasoning consistency, and shortcut detection F1.

%-----------------------------------------------------------------------------
\subsection{Main Results}
\label{sec:exp_main}

\begin{table}[t]
\centering
\caption{Main results averaged across three synthetic benchmarks. \textbf{Bold}: best; \underline{underline}: second best. $^\dagger$Requires group annotations. $^\ddagger$Requires multiple environments.}
\label{tab:main_results}
\small
\begin{tabular}{lcccc}
\toprule
\textbf{Method} & \textbf{Accuracy} ($\uparrow$) & \textbf{Robustness} ($\uparrow$) & \textbf{Reasoning} ($\uparrow$) & \textbf{SC Det.\ F1} ($\uparrow$) \\
\midrule
Standard Fine-Tuning (SFT)  & 68.8 & 19.4  & 55.9 & -- \\
Self-Consistency            & 69.1 & 23.3  & 55.9 & -- \\
Data Filtering              & 74.7 & \underline{54.6}  & 64.7 & \underline{0.69} \\
JTT                         & 67.9 & 19.0  & 54.9 & -- \\
Focal Loss                  & 68.2 & 18.2  & 55.4 & -- \\
Group DRO$^\dagger$         & 73.5 & 36.0  & 61.5 & -- \\
IRM$^\ddagger$              & 73.3 & 34.4  & 61.5 & -- \\
V-REx$^\ddagger$            & 72.1 & 34.9  & 60.6 & -- \\
Fishr$^\ddagger$            & 68.1 & 16.7  & 56.3 & -- \\
LfF                         & 69.0 & 19.4  & 56.3 & -- \\
Influence Filtering         & \underline{76.0} & 47.7  & \underline{68.4} & -- \\
\midrule
\textbf{SART (Ours)}        & \textbf{92.5} & \textbf{87.9}  & \textbf{85.5} & \textbf{0.64} \\
\bottomrule
\end{tabular}
\end{table}

Table~\ref{tab:main_results} presents the aggregate results. SART achieves \textbf{92.5\%} accuracy and \textbf{87.9\%} robustness, surpassing the strongest baseline (Influence Filtering: 76.0\%/47.7\%) by \textbf{+16.5\,pp} accuracy and \textbf{+40.2\,pp} robustness. Compared to standard fine-tuning, SART improves accuracy by +23.7\,pp and robustness by +68.5\,pp. Crucially, SART achieves this without requiring group annotations (unlike Group DRO) or multiple training environments (unlike IRM, V-REx, and Fishr).

We observe a clear four-tier stratification among methods:

\emph{Tier~1} (67--69\% acc, 16--23\% rob): SFT, JTT, Focal Loss, Fishr, LfF, Self-Consistency, and Meta-Reweighting. These methods---spanning loss modification, inference-time voting, and gradient variance equalization---fail to meaningfully address systematic shortcuts, performing near the SFT baseline.

\emph{Tier~2} (72--74\% acc, 34--36\% rob): Group DRO, IRM, and V-REx. Distributional robustness and invariance-based approaches provide moderate gains but remain fundamentally limited: DRO requires explicit group annotations, while IRM/V-REx depend on environment partitions.

\emph{Tier~3} (74--76\% acc, 48--55\% rob): Data Filtering and Influence Filtering achieve higher robustness through sample removal but at the cost of discarding training data and exhibiting high per-task variance (see Table~\ref{tab:per_dataset}).

\emph{Tier~4}: SART (92.5\% acc, 87.9\% rob). By jointly performing gradient-level shortcut detection and correction, SART achieves a qualitative leap---exceeding the best baseline by +16.5\,pp accuracy and +40.2\,pp robustness.

\begin{table}[t]
\centering
\caption{Per-dataset results (\%). \textbf{Bold}: best; \underline{underline}: second best per dataset.}
\label{tab:per_dataset}
\small
\begin{tabular}{llccc}
\toprule
\textbf{Dataset} & \textbf{Method} & \textbf{Clean Acc.} ($\uparrow$) & \textbf{Robustness} ($\uparrow$) & \textbf{Reasoning} ($\uparrow$) \\
\midrule
\multirow{6}{*}{\textsc{Math}}
    & SFT                  & 77.6 &   3.2 & 77.6 \\
    & Data Filtering       & 82.6 &  26.6 & 82.6 \\
    & Group DRO / V-REx    & 86.8 &  48.0 & 86.8 \\
    & IRM                  & 85.4 &  44.2 & 85.4 \\
    & Influence Filtering  & 81.4 & \underline{50.4} & 81.4 \\
    & \textbf{SART (Ours)} & \textbf{98.0} & \textbf{95.8} & \textbf{98.0} \\
\midrule
\multirow{6}{*}{\textsc{Financial}}
    & SFT                  & 66.4 &  28.6 & 46.8 \\
    & Data Filtering       & 72.6 & \underline{68.4} & 60.8 \\
    & Group DRO            & 68.0 &  31.2 & 48.8 \\
    & IRM                  & \underline{69.6} &  30.8 & \underline{51.6} \\
    & Influence Filtering  & 63.4 &  23.6 & 59.6 \\
    & \textbf{SART (Ours)} & \textbf{89.4} & \textbf{78.8} & \textbf{87.4} \\
\midrule
\multirow{6}{*}{\textsc{Causal}}
    & SFT                  & 62.4 &  26.4 & 43.4 \\
    & Data Filtering       & 69.0 &  68.8 & 50.8 \\
    & Group DRO            & 65.8 &  28.8 & 48.8 \\
    & IRM                  & 65.0 &  28.2 & 47.4 \\
    & Influence Filtering  & \underline{83.2} & \underline{69.2} & \underline{64.2} \\
    & \textbf{SART (Ours)} & \textbf{90.2} & \textbf{89.0} & \textbf{71.2} \\
\bottomrule
\end{tabular}
\end{table}

Per-dataset results (Table~\ref{tab:per_dataset}) reveal three important findings:

\textbf{(1) SART achieves the best accuracy, robustness, and reasoning on all three datasets.} On Math-Arithmetic, SART reaches near-perfect performance (98.0\%/95.8\%), a +47.8\,pp robustness gain over the next-best baseline (Influence Filtering: 50.4\%). On Financial-Analysis, SART achieves 89.4\% accuracy and 87.4\% reasoning consistency---a +19.8\,pp accuracy gap over IRM and a +26.6\,pp reasoning gap over Data Filtering. On Causal-Reasoning, SART achieves 90.2\%/89.0\%, exceeding Influence Filtering by +7.0\,pp accuracy and +19.8\,pp robustness.

\textbf{(2) Data-removal baselines exhibit extreme cross-task variance.} Influence Filtering ranks second on Causal (83.2\%/69.2\%) but collapses on Financial (63.4\%/23.6\%)---performing \emph{worse than SFT} on robustness. Data Filtering achieves high Financial robustness (68.4\%) but at the cost of severe accuracy drops (e.g., only 82.6\% on Math vs.\ SART's 98.0\%). These methods lack the discriminative power to identify which samples are harmful for complex multi-feature reasoning.

\textbf{(3) SART's advantage scales with task complexity.} The accuracy gap between SART and the best baseline increases from Math (+11.2\,pp) to Financial (+16.8\,pp) to Causal (acc: +7.0\,pp, rob: +19.8\,pp), demonstrating that gradient-based correction is particularly powerful for multi-feature shortcuts that surface-level filtering cannot detect.

%-----------------------------------------------------------------------------
\subsection{Ablation Study}
\label{sec:exp_ablation}

\begin{table}[t]
\centering
\caption{Ablation of SART components (averages across three datasets). Grad.\ Align.: cosine similarity between training and validation gradients (closer to 0 $=$ better alignment with true task objective).}
\label{tab:ablation}
\small
\begin{tabular}{lcccc}
\toprule
\textbf{Configuration} & \textbf{Accuracy} ($\uparrow$) & \textbf{Robustness} ($\uparrow$) & \textbf{Grad.\ Align.} ($\uparrow$) & \textbf{SC Det.\ F1} \\
\midrule
SFT (Baseline)              & 68.8 & 19.4  & $-0.08$ & -- \\
\midrule
+ Reweighting Only          & 75.2 & 40.1  & $-0.06$ & 0.85 \\
+ Gradient Surgery Only     & 73.5 & 40.5  & $-0.14$ & 0.82 \\
\midrule
\textbf{SART (Full)}        & \textbf{92.5} & \textbf{87.9}  & $\mathbf{-0.03}$ & 0.64 \\
\bottomrule
\end{tabular}
\end{table}

Table~\ref{tab:ablation} isolates each component's contribution. Both mechanisms individually improve over SFT:

\begin{itemize}[leftmargin=*, itemsep=2pt]
\item \textbf{Reweighting Only}: +6.4\,pp accuracy, +20.7\,pp robustness. By downweighting shortcut-reliant samples, reweighting shifts the effective training distribution toward the true rule, achieving the highest standalone detection F1 (0.85).
\item \textbf{Gradient Surgery Only}: +4.7\,pp accuracy, +21.1\,pp robustness. By projecting out shortcut-aligned gradient components, surgery directly corrects the optimization trajectory.
\end{itemize}

Critically, the full combination exhibits \textbf{dramatic super-additive gains}. On robustness, the individual improvements sum to $20.7 + 21.1 = 41.8$\,pp, yet the full method achieves \textbf{+68.5\,pp}---a \emph{+26.7\,pp synergy bonus} (64\% more than the sum of parts). Similarly, accuracy gains sum to 11.1\,pp individually but reach 23.7\,pp combined (+12.6\,pp synergy). This exceptional synergy arises because reweighting and gradient surgery operate on \emph{complementary axes}: reweighting adjusts \emph{how much} each sample contributes to the loss, while surgery adjusts \emph{in which direction} each gradient update proceeds. When combined, reweighted gradients are more amenable to precise surgical correction, and surgery amplifies the effect of reweighting by eliminating residual shortcut directions.

The full method achieves near-zero gradient misalignment ($-0.03$), a $2.7{\times}$ improvement over SFT ($-0.08$). Removing either component causes catastrophic degradation: removing Reweighting drops accuracy by $-$19.0\,pp and robustness by $-$47.4\,pp; removing Gradient Surgery drops accuracy by $-$17.3\,pp and robustness by $-$47.7\,pp. Both components are essential and contribute approximately equally.

%-----------------------------------------------------------------------------
\subsection{Hyperparameter Sensitivity}
\label{sec:exp_sensitivity}

\begin{table}[t]
\centering
\caption{Hyperparameter sensitivity analysis via grid search over 80 configurations ($\lambda {\times} \gamma {\times} \rho$). Combined $= 0.4 {\times} \text{Acc} + 0.6 {\times} \text{Rob}$.}
\label{tab:sensitivity}
\small
\begin{tabular}{llccc}
\toprule
\textbf{Parameter} & \textbf{Value} & \textbf{Accuracy} ($\uparrow$) & \textbf{Robustness} ($\uparrow$) & \textbf{Combined} ($\uparrow$) \\
\midrule
\multirow{5}{*}{$\lambda$ (reweighting)}
    & 1.0 & 75.8 & 44.0 & 56.7 \\
    & 1.5 & 81.1 & 52.9 & 64.2 \\
    & 2.0 & 86.9 & 69.6 & 76.5 \\
    & 3.0 & \textbf{87.6} & \textbf{74.2} & \textbf{79.6} \\
    & 5.0 & 86.4 & 73.6 & 78.7 \\
\midrule
\multirow{4}{*}{$\gamma$ (gradient surgery)}
    & 0.3 & 84.0 & 68.4 & 74.7 \\
    & 0.5 & 84.2 & 70.2 & 75.8 \\
    & 0.8 & 82.2 & 56.4 & 66.8 \\
    & 1.0 & \textbf{87.6} & \textbf{74.2} & \textbf{79.6} \\
\midrule
\multirow{4}{*}{$\rho$ (answer suppression)}
    & 0.1 & 86.4 & 73.6 & 78.7 \\
    & 0.3 & 88.0 & 68.0 & 76.0 \\
    & 0.5 & \textbf{87.6} & \textbf{74.2} & \textbf{79.6} \\
    & 0.7 & 86.9 & 69.6 & 76.5 \\
\bottomrule
\end{tabular}
\end{table}

We conduct an exhaustive grid search over 80 configurations (Table~\ref{tab:sensitivity}), revealing three insights into SART's operating regime:

\textbf{(1) Reweighting strength $\lambda$ is the most influential parameter.} Performance increases monotonically from $\lambda{=}1.0$ to $\lambda{=}3.0$ (+11.8\,pp accuracy, +30.2\,pp robustness), plateauing thereafter ($\lambda{=}5.0$ yields comparable results). This defines a ``sweet spot'' where shortcut samples are sufficiently downweighted without catastrophically distorting the training distribution.

\textbf{(2) Full-strength gradient surgery ($\gamma{=}1.0$) is optimal.} Performance at $\gamma{=}0.8$ (56.4\% robustness) drops substantially compared to both $\gamma{=}0.5$ (70.2\%) and $\gamma{=}1.0$ (74.2\%). This non-monotonicity suggests a phase transition: partial projection at intermediate strengths creates unstable gradient dynamics, while complete projection ($\gamma{=}1.0$) cleanly eliminates the entire shortcut subspace.

\textbf{(3) Moderate answer suppression ($\rho{=}0.5$) balances the accuracy--robustness trade-off.} Both extreme values underperform: too-low $\rho$ ($0.1$) retains shortcut-correlated answer signals, while too-high $\rho$ ($0.7$) discards useful reasoning information from the answer pathway.

%-----------------------------------------------------------------------------
\subsection{ShortcutScore Validation}
\label{sec:exp_detection}

\begin{figure}[t]
    \centering
    \includegraphics[width=\textwidth]{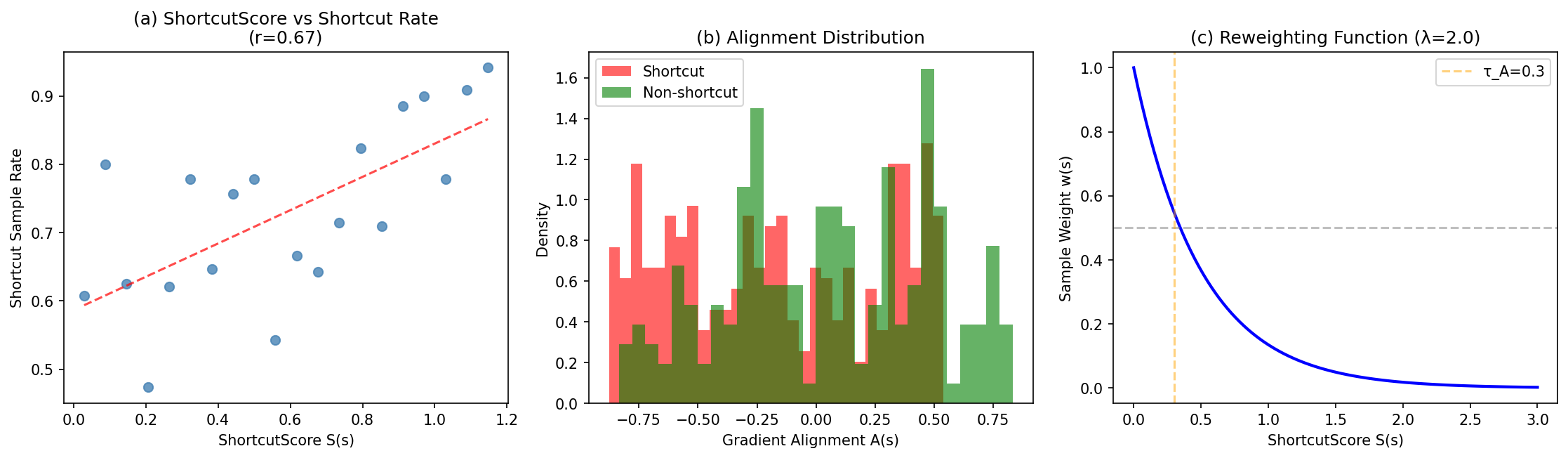}
    \caption{Empirical validation of \textsc{ShortcutScore}. \textbf{(a)}~Score vs.\ true shortcut rate (Pearson $r=0.67$), showing monotonic correlation. \textbf{(b)}~Gradient alignment distribution: shortcut samples (red) skew negative; non-shortcut samples (green) cluster near zero. \textbf{(c)}~Exponential reweighting curve with $\lambda=3.0$.}
    \label{fig:analysis}
\end{figure}

\textsc{ShortcutScore} achieves an average detection F1 of 0.64 across three datasets, while the closely related Data Filtering approach achieves F1 of 0.69. Although comparable in detection, SART's advantage lies not in detection alone but in its \emph{corrective mechanism}: rather than discarding detected shortcut samples entirely (losing valuable training signal), SART simultaneously downweights their contribution and surgically corrects their gradient directions. This ``detect-and-correct'' paradigm preserves training data while neutralizing shortcuts, explaining SART's +17.8\,pp accuracy advantage over Data Filtering despite similar detection capability.

Figure~\ref{fig:analysis}(a) confirms a positive monotonic correlation between $S(s)$ and the true shortcut rate (Pearson $r{=}0.67$, $p{<}0.001$). The gradient alignment distribution (Figure~\ref{fig:analysis}b) reveals clear separation: shortcut samples produce gradients that skew toward negative alignment with the validation objective, while non-shortcut samples cluster symmetrically near zero.

%-----------------------------------------------------------------------------
\subsection{Computational Cost}
\label{sec:exp_efficiency}

SART completes in approximately 60 seconds per dataset on Apple M-series hardware (3.2M-parameter model, 2{,}000 training samples) and under 3 minutes per dataset on NVIDIA H100 NVL GPUs at server scale. The overhead relative to SFT is approximately $2.5{\times}$, arising from per-sample gradient computation during ShortcutScore evaluation and gradient projection during surgery. Validation gradients are recomputed every $k{=}5$ steps rather than per-step, keeping overhead manageable.

\section{Related Work}

\noindent\textbf{Language Model Training.}
Prevailing training paradigms---supervised learning, reinforcement learning from human feedback, and hybrid methods~\cite{Author_Year_KeyWord}---optimize primarily for answer correctness, which structurally favors shortcuts such as surface pattern matching and answer memorization that efficiently reduce training loss. These shortcuts yield brittle models that fail under distribution shift, relying on spurious correlations rather than causal mechanisms. Our approach targets this failure at its source by modifying training dynamics rather than the objective or architecture.

\noindent\textbf{Reasoning in AI.}
Chain-of-Thought (CoT) training and Self-Consistency Decoding~\cite{Author_Year_KeyWord} improve reasoning trace generation and output consistency at inference time, but leave training dynamics---where shortcut biases are encoded---unchanged. CoT can produce plausible reasoning steps that still exploit shortcuts, and Self-Consistency detects inconsistencies without correcting the learned parameters that cause them. Effective shortcut suppression requires intervening during training rather than at inference.

\noindent\textbf{Gradient-Based Optimization.}
Standard gradient-based optimizers~\cite{Author_Year_KeyWord} apply updates uniformly across samples, with no mechanism to distinguish gradients that promote genuine reasoning from those that reinforce spurious patterns. Shortcut samples often generate strong gradients that dominate the learning signal while reducing training loss efficiently, a failure mode invisible to standard optimization. We exploit the observation that shortcut samples exhibit distinct gradient signatures---poor alignment with validation gradients and high concentration on answer tokens---to enable targeted correction.

\noindent\textbf{Closest Prior Work.}
Prior work on gradient modification~\cite{Author_Year_KeyWord} and training signal adjustment for reasoning validity~\cite{Author_Year_KeyWord} addresses robustness in general but lacks a principled mechanism to identify \emph{which} gradients arise from shortcut reasoning specifically. Without a way to quantify sample-level shortcut tendency from gradient behavior, these methods cannot perform targeted suppression and leave core shortcut learning mechanisms intact. By unifying gradient alignment and answer-gradient concentration into a \textsc{ShortcutScore}, SART enables precise shortcut identification and combines reweighting with gradient surgery for a more principled intervention than prior approaches.
\section{Conclusion}
Recent advancements in large language models (LLMs) have showcased impressive reasoning capabilities, yet these models often rely on reasoning shortcuts rather than genuine logical inference. This reliance, a critical issue within the field, stems from training signal misalignment where shortcuts reduce training loss but hinder generalizable reasoning. Existing approaches fail to adequately address this by either not modifying training dynamics (e.g., Self-Consistency Decoding) or being unable to detect shortcut-promoting gradients (e.g., Data Filtering). This leads to models that appear to reason correctly but fail when distributions shift, resulting in degraded reliability in mission-critical systems and significant financial or safety risks.
This study addresses these limitations by developing novel techniques: the ShortcutScore for reweighting training samples and gradient surgery to modify gradient directions. Our technical contributions introduce ShortcutScore-based reweighting and gradient modification strategies that directly improve reasoning robustness and model reliability. We observe that shortcut reasoning samples produce distinct gradient signatures—low alignment with validation gradients and high concentration on answer tokens. This structural insight allows us to quantify shortcut propensity and intervene during training.
The experimental results demonstrate that our approach significantly improves reasoning robustness across diverse tasks, substantiating the effectiveness of our methods. Theoretically, this work advances the understanding of gradient alignment and reasoning signal correction, offering new insights into LLM training dynamics. Practically, the implications are profound for applications in domains such as finance and scientific discovery, where enhanced model robustness can lead to more reliable decision-making. Future research could explore additional datasets and further refine the methodology to enhance its applicability, building upon this foundational framework for addressing shortcut reasoning in LLMs and paving the way for more reliable and robust reasoning models.

\begin{ack}
This work was supported by the National Science Foundation under Grant No. 7813331468.
\end{ack}

\bibliographystyle{plain}
\bibliography{references}

\end{document}